\documentclass[conference]{IEEEtran}
\IEEEoverridecommandlockouts
\usepackage{cite}
\usepackage{amsmath,amssymb,amsfonts}
\usepackage{algorithmic}
\usepackage{graphicx}
\usepackage{textcomp}
\usepackage{xcolor}
\usepackage{epsfig}
\usepackage{graphicx}
\usepackage{multirow}

\def\BibTeX{{\rm B\kern-.05em{\sc i\kern-.025em b}\kern-.08em
    T\kern-.1667em\lower.7ex\hbox{E}\kern-.125emX}}

\begin{document}

\title{FMHash: Deep Hashing of In-Air-Handwriting for User Identification
}

\author{Duo Lu, Dijiang Huang, Anshul Rai\\
Arizona State University, Tempe, Arizona\\
{\tt\small {\{duolu, dijiang.huang, anshulrai\}}@asu.edu}
}


\maketitle

\begin{abstract}
Many mobile systems and wearable devices, such as Virtual Reality (VR) or Augmented Reality (AR) headsets, lack a keyboard or touchscreen to type an ID and password for signing into a virtual website. However, they are usually equipped with gesture capture interfaces to allow the user to interact with the system directly with hand gestures. Although gesture-based authentication has been well-studied, less attention is paid to the gesture-based user identification problem, which is essentially an input method of account ID and an efficient searching and indexing method of a database of gesture signals. In this paper, we propose FMHash (\textit{i.e.}, Finger Motion Hash), a user identification framework that can generate a compact binary hash code from a piece of in-air-handwriting of an ID string. This hash code enables indexing and fast search of a large account database using the in-air-handwriting by a hash table. To demonstrate the effectiveness of the framework, we implemented a prototype and achieved $\ge$99.5\% precision and $\ge$92.6\% recall with exact hash code match on a dataset of 200 accounts collected by us. The ability of hashing in-air-handwriting pattern to binary code can be used to achieve convenient sign-in and sign-up with in-air-handwriting gesture ID on future mobile and wearable systems connected to the Internet.
\end{abstract}


\section{INTRODUCTION}

Gesture user interfaces are considered as the future way for people to interact with Virtual Reality (VR) or Augmented Reality (AR) applications \cite{HoloLens} and other devices such as a smart door bell or smart watch \cite{Soli}. Such interfaces can capture and track hand motions in the air, and allow a user to manipulate menus, dialogues, and other virtual objects directly by hand gesture. However, for security related tasks such as sign-up and sign-in, entering the user ID string and password through a virtual keyboard on gesture interfaces become cumbersome due to the lack of keystroke feedback. Existing researches \cite{ArmSweep, Feasibility, uWave, Madrid-Analysis, LeapPassword, KinWrite, FMCode-IJCB, FMCode-ICB} exploit the rich information in native gestures, and esp., in-air-handwriting, to authenticate a user. Yet, a usually neglected function is user identification. Authentication is a true or false question, \textit{i.e.}, answering whether the user owns the account which he or she claims to own. On the other hand, identification is a multiple choice question, \textit{i.e.}, answering which account the user wants to login among a database of many accounts. If we make an analogy of the sign-in procedure on a web with a desktop computer, the authentication procedure resembles typing and checking the password, while the identification procedure resembles searching the database given an ID number or ID string. Is it possible to construct a system that is capable of (1) taking a piece of in-air-handwriting of an ID string instead of typing, (2) searching a potentially large database of accounts registered using the in-air-handwriting, and (3) returning the matched identity or account number with high accuracy and short respond time?

There are challenges for gesture-based user identification due to the unique characteristics of the hand motion. First, hand motion has inherent fuzziness. Even if the same user writes the same string in the air twice, the generated two motion signals are not identical, but with minor variations. Yet, the system should be able to tolerate the fuzziness and identify these two signals as the same user. This is different from typing an ID string of characters twice where even a single bit difference in the typed ID can cause failure in the identification. Second, it is difficult for many native gestures to provide enough information to enable a large account ID space as well as distinctiveness. Third, the traditional method of hash table for indexing a large account database using an ID string or account number does not work with handwriting signal, unless there is a way to generate a fixed size binary hash code from the signal with tolerance of inherent fuzziness (\textit{i.e.}, fuzzy hash). 

\begin{figure*}
\begin{center}
\includegraphics[width=7in]{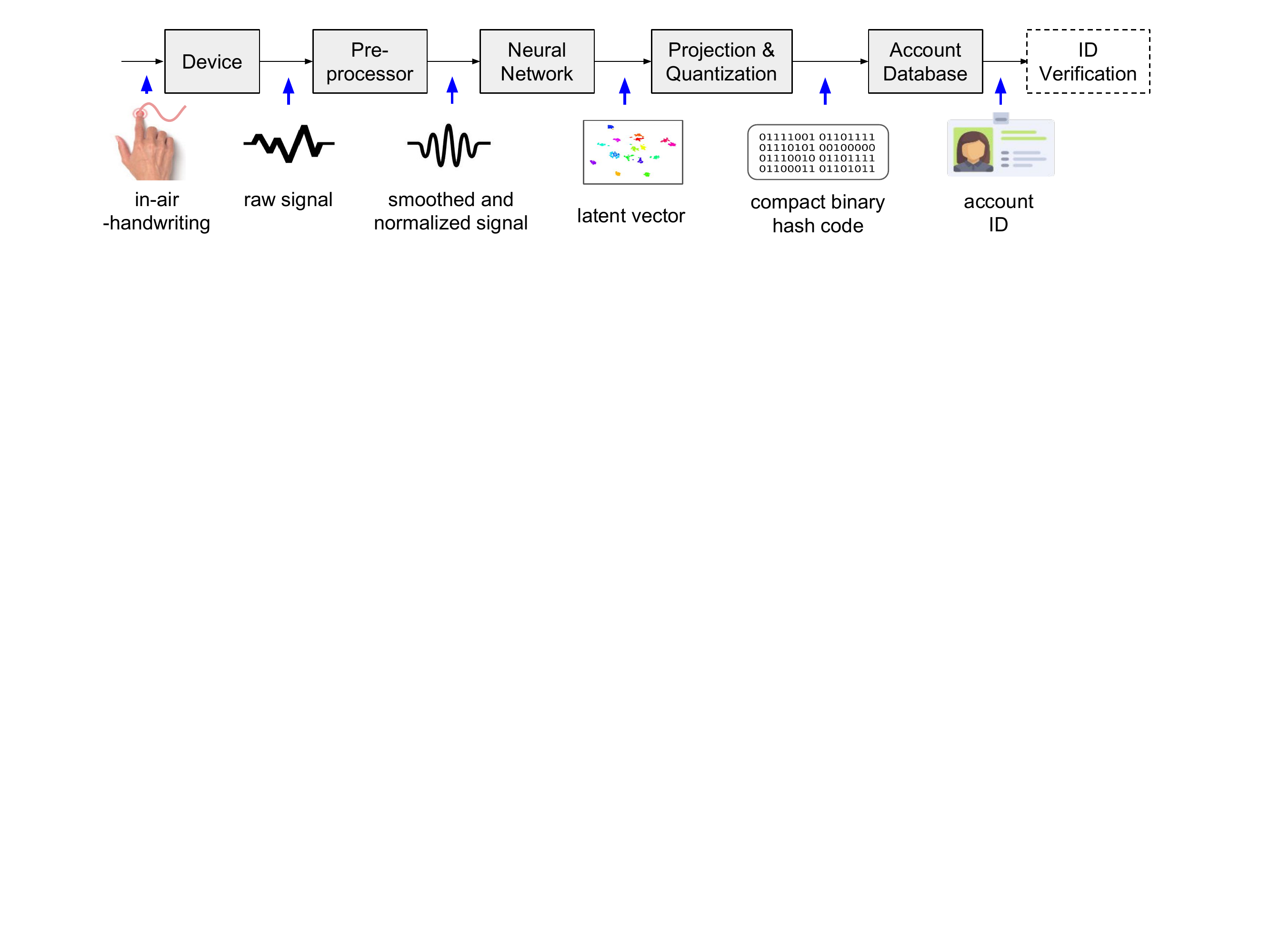}
\vspace{-4.1in}
\end{center}
   \caption{FMHash Framework Architecture.}
\label{fig:arch}
\vspace{-0.1in}
\end{figure*}

In this paper, we propose a framework called \textbf{FMHash}, i.e., Finger Motion Hash, to efficiently obtain a user's account ID from the hand motion signal of writing an ID string in the air. FMHash uses a camera-based gesture user input device to capture the hand motion and a deep convolutional neural network (called FMHashNet) to convert the in-air-handwriting signal to a binary hash code (\textit{i.e.}, deep hashing). With the hash code of the signals of all accounts, it further builds a hash table to index the whole account database to enable efficient user identification with hash table search. This is similar to face recognition, where a large database of identities are indexed using faces and an ID can be retrieved by presenting an image of a face. However, FMHash has a few unique advantages compared to face recognition. For example, one face is linked to one person, and a user can neither have multiple faces for multiple accounts nor change or revoke his or her own face. Moreover, the users may be worried about privacy because it is impossible to stay completely anonymous if a website requires face image to register. Yet, with in-air-handwriting of an ID string, the user can have multiple accounts with different ID strings, change or revoke the ID string, and stay anonymous by writing something unrelated to the true identity. In summary, our contributions in this paper are as follows:

\textbf{1)} We proposed a deep hashing framework of in-air-handwriting for user identification over gesture interface. To our best knowledge, FMHash is the first framework that can generate fuzzy hash code from in-air-handwriting. Our method can accommodate gesture fuzziness by hashing multiple instances of the same handwriting by the same user to the same binary code with high probability of success ($\ge$99.5\% precision and $\ge$92.6\% recall with exact hash code match).

\textbf{2)} We designed a regularizer called the \textit{pq-regularizer} and a progressive training procedure for our neural network model. With the \textit{pq-regularizer}, hashcode of in-air-handwriting signals of different accounts are separated more than two bits in over 99\% of the change. Meanwhile, we can maintain a reasonably fast training speed ($\sim$10 minutes for a full training on our dataset).

\textbf{3)} We provided a detailed analysis on the hash code fuzziness with a dataset of 200 accounts collected by us.

The remainder of the paper is organized as follows. Related works on gesture-based authentication and deep hashing are discussed in section II. In section III, the architecture of the proposed framework is presented. Then we show the empirical evaluation results in section IV. Finally we draw the conclusion and discuss future work in section V.

\section{RELATED WORKS}

Most 3D hand gesture based user authentication systems use a combination of password and behavioral biometrics, \textit{i.e.}, different users are differentiated in the gesture content or the convention of hand motion (sometimes both). The hand motion can be captured by a handheld device \cite{uWave, Madrid-Analysis}, wearable device \cite{ArmSweep, FMCode-IJCB}, or a camera \cite{LeapPassword, KinWrite, FMCode-ICB}. The authentication system compares the captured motion signal with a template \cite{uWave, KinWrite, FMCode-ICB} or runs a pipeline of feature extractors and statistical pattern classifiers \cite{Madrid-Analysis, FMCode-IJCB} to make a decision. The gesture content can be simple movements like shaking \cite{ArmSweep} or complex in-air-handwriting of a password or signature \cite{LeapPassword, KinWrite, MoCRA}. As mentioned previously, identification is different from authentication. Existing systems need to exhaustively check every account in the database using the authentication procedure, \textit{i.e.}, in $O(N)$ complexity, which is impractical with a large account database. Instead, we take a different route by converting a in-air-handwriting signal to a binary code using a deep hashing method for efficient user identification, \textit{i.e.}, in $O(1)$ complexity.

Deep hashing has been investigated in communities of computer vision and machine learning for image retrieval \cite{xia2014supervised, lai2015simultaneous, zhao2015deep, zhu2016deep, DSH, lin2016learning, cao2017deep, Liu_2017_CVPR} instead of security. Such image searching systems train a deep convolutional neural network (CNN) to convert 2D images to compact binary hash codes, and use the hash code to index a database with millions of images. To search similar images, a query image is converted to a hash code with the same neural network, and images in the database with similar hash code (\textit{i.e.}, in Hamming distance) are returned. Training such a neural network requires techniques of pairwise supervision \cite{xia2014supervised, DSH}, triplet supervision \cite{zhang2015bit, lai2015simultaneous}, or various careful design of regularization and quantization \cite{DSH}, as well as special treatment of the image like salience mask \cite{jin2018deep}. Although we are also utilizing CNN to generate hash code, our work has differences. First, the features of in-air-handwriting motion signal are fundamentally different from the features of an image. Second, user identification has different goals from image retrieval. Our identification system desires sparsity in the hash code space to avoid wrong identification because each ID is unique (using the \textit{pq-regularizer}); while an image searching system desires smooth similarity change of the hash code to cope with the smooth semantic shift of images. Meanwhile, long hash code is preferred in identification to defend random guess. Third, considering that new accounts are registered from time to time, a neural network for user identification must be optimized with smaller number of parameters and faster training speed so as to be retrained online frequently.

It should also be noted that FMHash is not a biometric verification system, and an ID string is not a signature linked to personal identity. An ID string can be created by a user containing arbitrary content as long as it has distinctiveness in the account database.

\section{THE FMHASH FRAMEWORK}

The proposed FMHash framework (shown in Fig. \ref{fig:arch}) contains six components: 

(1) An in-air-handwriting motion capture device (e.g., a Leap Motion controller \cite{LeapMotion} in our implementation);

(2) A preprocessing module smoothing and normalizing the captured motion signal (detailed in section III.A);

(3) A deep neural network that takes preprocessed motion signal $\mathbf{x}$ as input and generate a high dimensional floating point latent vector $\mathbf{h}$ (denoted as a function $f(\mathbf{x})=\mathbf{h}$, detailed in section III.B);

(4) An additional neural network layer that projects the latent vector $\mathbf{h}$ to low dimensional space and quantize the projected result to $B$-bit binary fuzzy hash code $\mathbf{b} \in \{-1, +1\}^B$ (denoted as another function $g(\mathbf{h})=\mathbf{b}$, and $B$ ban be 16, 32, 64, etc., also detailed in section III.B); 

(5) An account database that stores a hash table index of account tuples $<$ID, $\mathbf{b}^{ID}$, $\mathbf{h}^{ID}>$, where ID is the account ID (usually a unique number generated by the system at registration), $\mathbf{b}^{ID}$ and $\mathbf{h}^{ID}$ are the hash code and latent vector corresponding to the account (detailed in section III.C);

(6) Optionally, there is an verification module after an ID is obtained by the FMHash framework. This ID is a candidate ID. The system can run a procedure similar to the authentication procedure by comparing the in-air-handwriting of the ID string to some information stored in the account referred by the candidate ID, which can further eliminate wrong identification results (detailed in section III.D).

\subsection{Signal Acquisition and Preprocessing}

\begin{figure}[t]
\begin{center}
\includegraphics[width=3.4in]{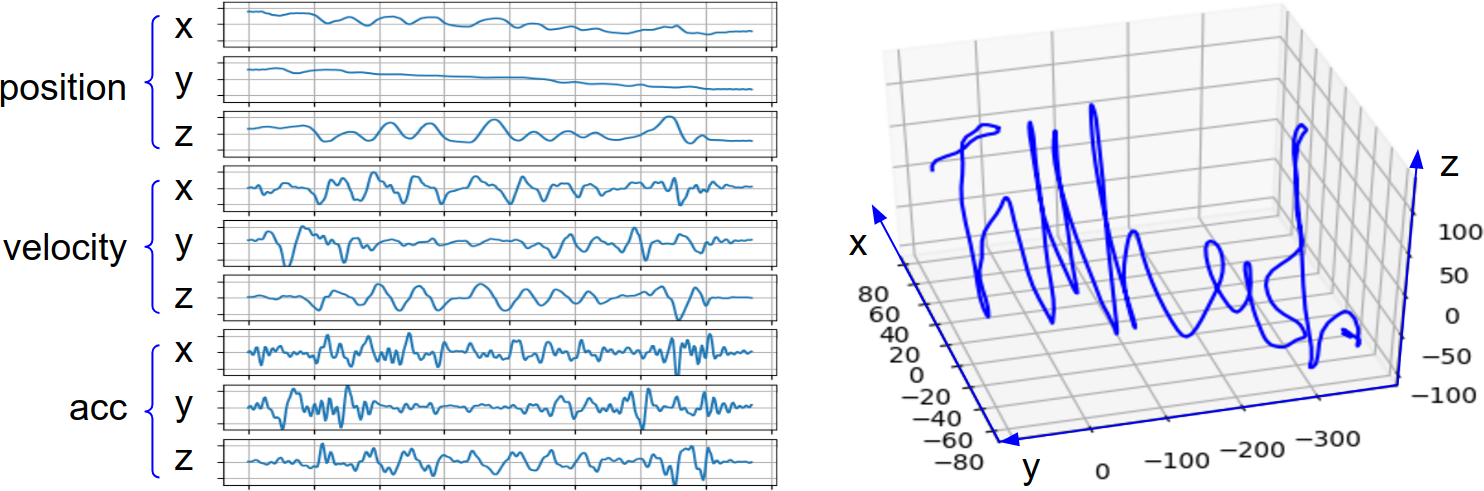}
\end{center}
   \caption{An example of the motion signal (left) and trajectory in the 3D space (right) obtained by writing ``FMhash'' in the air.}
\label{fig:sig_exp}
\end{figure}

The in-air-handwriting of an ID string is captured by the Leap Motion controller in our implementation as a raw signal containing the 3D position coordinates of the center of the hand sampled at about 110 Hz. Once this raw signal is obtained, we further extract the 3D velocity, and 3D acceleration using the difference of adjacent position samples. Then the signal is normalized in hand pose (making the average hand pointing direction as x-axis) and amplitude (mean subtraction and division by standard deviation). Finally it is resampled to a fixed length of 256 data points in each dimension to form the 256$\times$9 input vector $\mathbf{x}$. An example of the motion signal and the trajectory of in-air-handwriting is shown in Fig. \ref{fig:sig_exp}.

\subsection{FMHashNet}

\begin{table}[]
\small
\begin{center}
\label{tb:nn_arch}
\begin{tabular}{|c|c|c|c|}
\hline
layer      & kernel                      & output   & \#para \\ \hline
\multicolumn{4}{|c|}{input: 256 * 9}                          \\ \hline
conv-pool1 & 3$\rightarrow$1 conv, 2$\rightarrow$1 max pool & 128 * 48 & 1.3k      \\ \hline
conv-pool2 & 3$\rightarrow$1 conv, 2$\rightarrow$1 max pool & 64 * 96  & 14k      \\ \hline
conv-pool3 & 3$\rightarrow$1 conv, 2$\rightarrow$1 max pool & 32 * 128  & 37k     \\ \hline
conv-pool4 & 3$\rightarrow$1 conv, 2$\rightarrow$1 avg pool & 16 * 192 & 74k     \\ \hline
conv-pool5 & 3$\rightarrow$1 conv, 2$\rightarrow$1 avg pool & 8 * 256  & 147k     \\ \hline
fc (latent) & fully connected & 512  & 1,048k     \\ \hline
\end{tabular}
\begin{tabular}{|c|c|c|c|c|c|c|}
\cline{1-3} \cline{5-7}
layer   & output & \#para &  & layer      & output & \#para  \\ \cline{1-3} \cline{5-7} 
softmax & 200    & 102k    &  & projection & B      & 512*B \\ \cline{1-3} \cline{5-7}
\multicolumn{3}{|c|}{cross-entropy loss} & \multicolumn{1}{l|}{} & \multicolumn{3}{c|}{pairwise loss} \\ \cline{1-3} \cline{5-7}
\end{tabular}
\end{center}
\caption{FMHashNet Architecture}
\vspace{-0.2in}
\end{table}

The deep neural network and the additional projection-quantization layer are implemented together, which are collectively called the FMHashNet. There are multiple design goals of this neural network. First, for a pair of in-air-handwriting signals $(\mathbf{x}_1, \mathbf{x}_2)$, if they are generated by the same user writing the same ID string, the corresponding hash codes $(\mathbf{b}_1, \mathbf{b}_2)$ should be the same in most cases or differ only in one or two bits sometimes due to the fuzziness of the signals. However, if they are generated from different ID strings (regardless of the same user or different users), $(\mathbf{b}_1, \mathbf{b}_2)$ should differ by at least three bits. Second, the neural network should learn contrastive representations $\mathbf{h}$ to facilitate the projection. Third, it should be easy to train and fast to converge.

To achieve these goals, we design the FMHashNet in the following way, as shown in Table 1. First, we apply five convolutional and pooling layers with VGG-like kernel \cite{VGG} and a fully connected layer to map input signal $\mathbf{x}$ to latent vectors $\mathbf{h}$. Both the convolutional layer and the fully connected layer have leaky ReLU activation. Next, the projection layer projects the latent vector $\mathbf{h}$ to a space in the same dimension as the final hash code, i.e., $\mathbf{z} = W\mathbf{h} + \mathbf{c}$, where $\mathbf{z}$ is the projected vector whose dimension is $B$. Here $W$ and $c$ are trainable parameters. After that, the hash code is generated by taking the sign of the projected vector $b_i = sign(z_i), 1 \le i \le B$. This is essentially partitioning the latent space by $B$ hyperplanes to obtain at most $2^B$ regions, where each region is associated with a unique hash code. Additionally, a softmax layer is added in parallel with the projection layer to help training the neural network.

Training the FMHashNet is equivalent to placing all registered accounts into these $2^B$ regions, which is achieved progressively in two steps. First, we train the network with the softmax layer and cross-entropy loss to allow the convolutional filters to converge. In this step the projection layer is not activated. Note that the softmax classification layer does not need to contain all accounts if the account number is large, or even a independent dataset for pretraining can be utilized. 

Second, we train the network using the projection layer with the following pairwise loss $L$, and minibatches of $2M$ pairs of signals $(\mathbf{x}^{(i)}_1, \mathbf{x}^{(i)}_2), 1 \le i \le 2M$ . Here $2M$ (\textit{i.e.}, the batch size) is a hyperparameter chosen empirically. Larger batch size leads to more computation in the training but more stable convergence behavior. In the minibatch, half pairs of signals are from the same account ($y^{(i)}$ = 0), and the other half pairs of signals are from different accounts ($y^{(i)}$ = 1). The loss function $L$ is as follows:
$$L = \frac{1}{2M}\sum_{i = 1}^{2M}L^{(i)},$$
$$L^{(i)} = (1 - y^{(i)}) ||\mathbf{z}_1^{(i)} - \mathbf{z}_2^{(i)}||
+ y^{(i)} \mbox{max}(m - ||\mathbf{z}_1^{(i)} - \mathbf{z}_2^{(i)}||, 0)$$
$$+ \alpha (P(\mathbf{z}_1^{(i)}) + P(\mathbf{z}_2^{(i)})) + \beta (Q(\mathbf{z}_1^{(i)}) + Q(\mathbf{z}_2^{(i)})).
$$
\noindent Here $||.||$ is the Euclidean norm. In this loss function, the first term forces the projected vectors of the same account to the same value, and the second term forces the projected vectors of different accounts to separate at least $m$ in Euclidean distance. The remaining terms $P(\mathbf{z})$ and $Q(\mathbf{z})$ are the so-called \textit{pq-regularizer} which is specially designed to help place all registered accounts into different regions and avoid ambiguity in quantization. These two terms are defined as follows:
$$ P(\mathbf{z}^{(i)}) = \sum_{j = 1}^{B} \mbox{max}(|z_j^{(i)}| - p, 0),$$
$$ Q(\mathbf{z}^{(i)}) = \sum_{j = 1}^{B} \mbox{max}(q - |z_j^{(i)}|, 0),$$
where $p$ and $q$ are hyperparameters chosen empirically, $|z_j^{(i)}|$ is taking absolute value of the $jth$ component of $\mathbf{z}^{(i)}$. This regularizer forces each element of the projected vector $\mathbf{z}$ to reside in the region $[-p, -q]$ or the region $[+q, +p]$. The element $\mathbf{z}^{(i)}$ is quantized to the bit -1 if it is less than zero or bit +1 if it is greater or equal to zero, so as to generate the final fuzzy hash code bit $b_i$. The $p$ prevents $\mathbf{z}^{(i)}$ taking unbounded large values, and the $q$ prevents $\mathbf{z}^{(i)}$ taking values close to zero which causes quantization ambiguity. 

\begin{figure}[t]
\begin{center}
\includegraphics[width=3.25in]{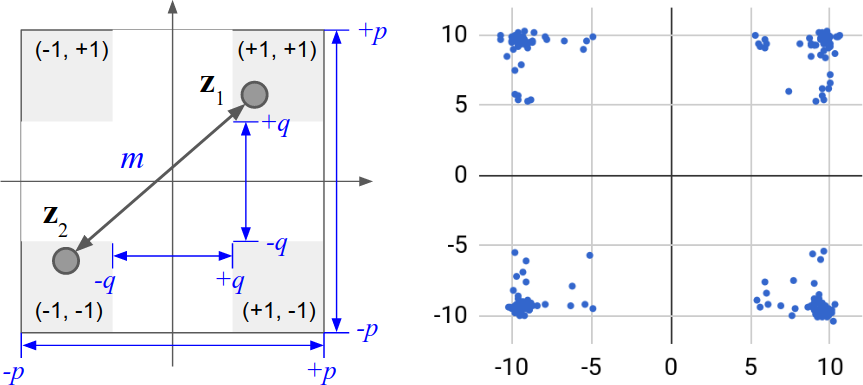}
\end{center}
   \caption{The effect of the pq-regularizer that pushes $(\mathbf{z}_1, \mathbf{z}_2)$ from different accounts to different regions, in illustration (left) and with actual data (right). The right figure is obtained by plotting the first two dimensions of $\mathbf{z}$ of 200 training signals from 200 different accounts, with p=10 and q=5.}
\label{fig:regularizer}
\vspace{-0.1in}
\end{figure}

With a careful choice of $m$, we can push a pair of $(\mathbf{z}^{(i)}_1, \mathbf{z}^{(i)}_2)$ of different accounts to opposite regions, and hence, hash them to different binary codes, as shown in Fig. \ref{fig:regularizer}. One example choice of $m$ as in our experiment is $p \sqrt{B}$, which is the Euclidean distance from the origin to the point $\mathbf{z}^\ast = (p, p, ..., p)$. This forces the hash code of signals of different accounts differ at least one bit. Our experience shows larger $m$ helps separation, but hurts convergence. The hyperparameter $\alpha$, $\beta$ controls the portion of contribution of the regularizer in the total loss and gradients. Our design differs from most related works of deep hashing which try to minimize quantization loss (i.e., forces the projected vector to be close to the vertices of Hamming hypercube). Instead, we map the input to a bounded Euclidean space and push them away from the decision boundary $z_j = 0$, where a relatively large region that can be quantized to the same bit value regardless of the quantization error. The effectiveness of our regularizer relies on the assumption that ID strings are distinctive, which is true in an identification system but not in an image retrival system. Meanwhile, both the FMHashNet activation function and the regularizer are piece-wise linear, which is easier to compute and train compared to the saturation methods such as tanh or sigmoid relaxation commonly used in deep hashing.

\subsection{Account Database}

As mentioned previously, each account contains a tuple of $<$ID, $\mathbf{b}^{ID}$, $\mathbf{h}^{ID}>$. At registration time, the system generates a unique account ID for the registered account. The user is asked to create an ID string and write it $K$ times. The obtained $K$ in-air-handwriting signals $\{\mathbf{x}^{(1)}, \mathbf{x}^{(2)}, ..., \mathbf{x}^{(K)}\}$ are utilized to train the FMHashNet. Once the training is finished, we can use the training signals to construct $\mathbf{b}^{ID}$ and $\mathbf{h}^{ID}$ as follows:
$$\mathbf{h}^{ID} = \frac{1}{K} \sum_{i = 1}^{K}\mathbf{h}^{(i)} = \frac{1}{K} \sum_{i = 1}^{K}f(\mathbf{x}^{(i)}),$$
$$\mathbf{b}^{ID} = g(\mathbf{h}^{ID}) = \mbox{sign}(W\mathbf{h}^{ID} + \mathbf{c}),$$
\noindent where $f()$ is the deep neural network $g()$ is the projection and quantization process, and $sign()$ is element-wise sign function. A hash table is also constructed to index all account tuples using the hash codes $\mathbf{b}^{ID}$.

At identification time, given a preprocessed in-air-handwriting signal $\mathbf{x}^\prime$, the following steps are proceeded to obtain the account ID. First, we run the forward path of FMHashNet to obtain a latent vector $\mathbf{h}^\prime$ and $\mathbf{b}^\prime$. Second, we search the hash table using $\mathbf{b}^\prime$ with a tolerance of $l$ bit. If $l$ is 0, we just search the hash table using $\mathbf{b}^\prime$. If $l$ is not 0, we search the hash table multiple times with each element of a collection of hash codes $S$, where $S$ contains all possible hash codes with a Hamming distance less or equal than $l$ bits from $\mathbf{b}^\prime$. The rationale is that the fuzziness in the writing behavior eventually lead to errors that make $\mathbf{b}^\prime$ differ from the hash code of its real account, but this difference should be smaller than $l$ bits. In practice, we usually set $l$ to 1 or 2 to limit the total number of searches for prompt response. In this way, a collection of candidate accounts will be obtained. The third step is compare $\mathbf{h}^\prime$ with the latent vector of every candidate account to find the nearest neighbor. Finally, the account ID of this nearest neighbor is returned as the identified ID.

\subsection{ID Verification}

In this identification procedure explained previously, the nearest neighbor search step serves as a verification of the ID. Alternatively, the system may store a template of the handwriting of the ID string generated at registration for each account, instead of the $\mathbf{h}^{ID}$. Upon an identification request, the system can compare the signal in the request with the templates of all candidate accounts obtained by the hash table search and run a procedure similar to an authentication system to verify the candidate IDs. The motivation is that the hashing step loses information in the in-air-handwriting, which may lead to collisions if an imposter writes the same ID string as a legitimate user, and hence, a verification step can reduce misidentification significantly. Since the verification step is essentially an authentication system, we will not elaborate and evaluate it in this paper. Besides, an attacker can create colliding signals if both the hash code $\mathbf{b}^{ID}$ and the parameters of the neural network are leaked because FMHashNet generates fuzzy hash codes instead of crypotographic hash code. In practice, $\mathbf{b}^{ID}$ can be hashed again using a crypotographic hash algorithm such as SHA-256 for the hash table, while searching with bit tolerance can still work ($S$ contains crypotographically hashed elements of the original $S$). In this case, the crypotographic hash of $\mathbf{b}^{ID}$ is stored in the account database. Moreover, this crypotographic hash can be further used to generate a key to encrypt the template for the ID verification to further improve the security.

\section{EXPERIMENTAL EVALUATION}

\subsection{Dataset}

To evaluate the FMHash system, we collected a dataset of 200 accounts with 200 distinct ID strings, created by 100 users with exactly two accounts per user. For each account, the user wrote an ID string five times as registration (\textit{i.e.}, training) and another five times as five independent identification requests (\textit{i.e.}, testing). Roughly half of the users are college students, and the other half are people of other various occupations (including both office workers and non-office workers). The contents of the ID strings are determined by the users and no two ID strings are identical. Most users chose a meaningful phrase so that it is easy to remember and they wrote the ID strings very fast in an illegible way for convenience. The average time of writing an ID string in the air is around 3 to 8 seconds.

\begin{figure}[t]
\begin{center}
\includegraphics[width=3in]{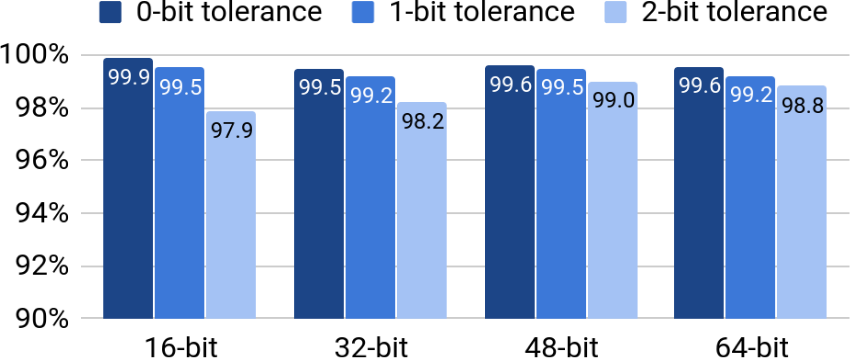}
\end{center}
\vspace{-0.1in}
   \caption{Average Precision}
\label{fig:precision}
\end{figure}

\begin{figure}[t]
\begin{center}
\includegraphics[width=3in]{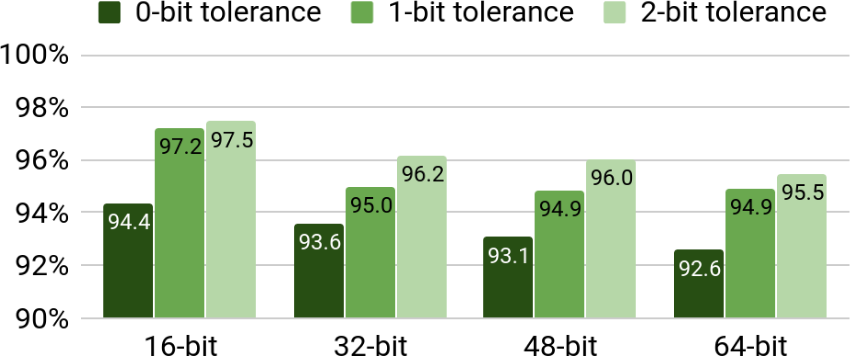}
\end{center}
\vspace{-0.1in}
   \caption{Average Recall}
\vspace{-0.2in}
\label{fig:recall}
\end{figure}

\begin{figure}[t]
\begin{center}
\includegraphics[width=3in]{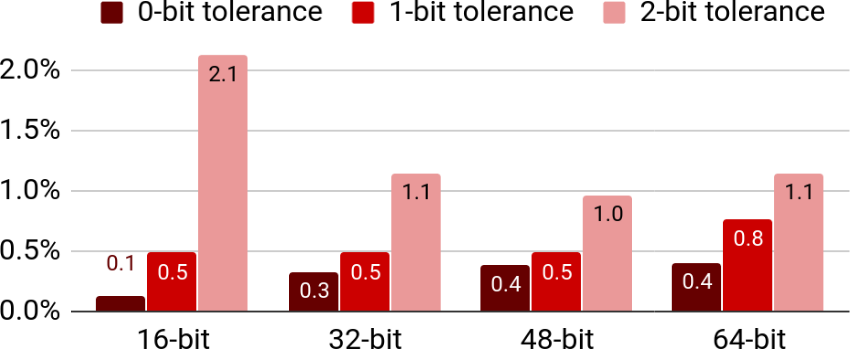}
\end{center}
\vspace{-0.1in}
   \caption{Misidentification Rate}
\label{fig:missrate}
\end{figure}

\begin{figure}[t]
\begin{center}
\includegraphics[width=3in]{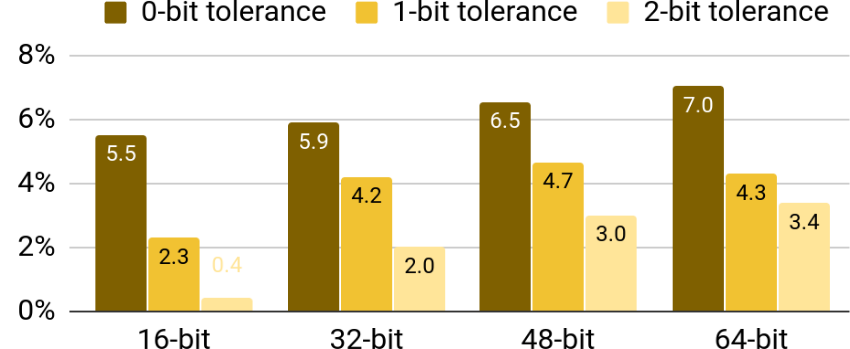}
\end{center}
\vspace{-0.1in}
   \caption{Failure of identification Rate}
\vspace{-0.2in}
\label{fig:failrate}
\end{figure}

\subsection{Implementation Details}

The FMHashNet is implemented in TensorFlow \cite{TensorFlow} on a Nvidia GTX 1080 Ti GPU. The weight parameters are initialized with the Xavier method \cite{Xavier} and the Adam optimizer \cite{Adam} with a initial learning rate of 0.001 is used. The leaky ReLU negative slope is set to 0.2. The regularizer hyperparameter $p$ is set to 10, $q$ is set to 5. The inter-class distance $m$ is set to $p \sqrt{B}$, and the hash code size $B$ is 16, 32, 48 and 64. For the training protocol, we first use the softmax layer and cross-entropy loss with 1,000 iterations. Then we use the projection layer and pairwise loss with pq-regularizer for another 10,000 iterations. During these 10,000 iterations, $\alpha$ is always set to 0.1, and $\beta$ is initially set 0.0001 for the first 2,000 iterations, and gradually increased 10 times per every 2,000 iterations until 0.1. The training pairs are selected online, and $M$ is set to 200 in a minibatch. For the pairs of the same account, we randomly select an account and two training signals of that account; for the pairs of different accounts, we calculate the account hash code $\mathbf{b}^{ID}$ for each account every 20 iterations, and select pairs from those accounts whose hash codes differs less than three bits. If no such account exists, we randomly choose two signals from two different accounts as a pair.

\begin{table*}[t]
\centering
\caption{Performance Comparison (with hash code side $B$ = 16)}
\small
\label{tb:compare}
\begin{tabular}{|l|r|r|r|r|r|r|r|r|r|r|r|r|l|}
\hline
\multirow{2}{*}{methods} & \multicolumn{3}{c|}{average precision}                                                                          & \multicolumn{3}{c|}{average recall}                                                                             & \multicolumn{3}{c|}{miss-rate}                                                                                  & \multicolumn{3}{c|}{fail-rate}                                                                                  & \multicolumn{1}{c|}{\multirow{2}{*}{\begin{tabular}[c]{@{}c@{}}training\\ time\end{tabular}}} \\ \cline{2-13}
                  & \multicolumn{1}{l|}{\textit{0 bit}} & \multicolumn{1}{l|}{\textit{1 bit}} & \multicolumn{1}{l|}{\textit{2 bit}} & \multicolumn{1}{l|}{\textit{0 bit}} & \multicolumn{1}{l|}{\textit{1 bit}} & \multicolumn{1}{l|}{\textit{2 bit}} & \multicolumn{1}{l|}{\textit{0 bit}} & \multicolumn{1}{l|}{\textit{1 bit}} & \multicolumn{1}{l|}{\textit{2 bit}} & \multicolumn{1}{l|}{\textit{0 bit}} & \multicolumn{1}{l|}{\textit{1 bit}} & \multicolumn{1}{l|}{\textit{2 bit}} & \multicolumn{1}{c|}{}                                                                         \\ \hline
DSH-like \cite{DSH}              & 0.995                               & 0.916                               & 0.636                               & 0.918                               & 0.892                               & 0.632                               & 0.004                               & 0.081                               & 0.362                               & 0.078                               & 0.026                               & 0.005                               & 648 s                                                                                         \\ \hline
FMHashNet (tanh)         & 0.970                               & 0.821                               & 0.494                               & 0.443                               & 0.638                               & 0.474                               & 0.014                               & 0.139                               & 0.484                               & 0.544                               & 0.223                               & 0.042                               & 637 s                                                                                         \\ \hline
FMHashNet              & \textbf{0.999}                      & \textbf{0.995}                      & \textbf{0.979}                      & \textbf{0.944}                      & \textbf{0.972}                      & \textbf{0.975}                      & \textbf{0.001}                      & \textbf{0.005}                      & \textbf{0.021}                      & \textbf{0.055}                      & \textbf{0.023}                      & \textbf{0.004}                      & 610 s                                                                                         \\ \hline
\end{tabular}
\end{table*}

Another major challenge we encountered is the limited amount of training data (only five signals per account). To overcome this challenge, we augment the training dataset in two steps. First, given $K$ signals $\{\mathbf{x}^{(1)}, \mathbf{x}^{(2)}, ..., \mathbf{x}^{(K)}\}$ obtained at registration, for each $\mathbf{x}^{(k)}$ in this set, we align all the other signals to $\mathbf{x}^{(k)}$ to create $K - 1$ additional signals using Dynamic Time Warping \cite{DTW}, and in total we can obtain $K^2$ signals (in our case 25 signals). Second, we randomly pick two aligned signals and exchange a random segment to create a new signal, and this step is repeated many times. Finally each account has 125 training signals.

\subsection{Empirical Results}

We trained the FMHashNet with different hash code sizes $B =$ 16, 32, 48, 64 and tested it with fuzziness tolerances $l =$ 0, 1, 2. In a single experiment, we train the FMHashNet from scratch with the 200$\times$125 augmented training signals and ran the identification procedure with the 200$\times$5 testing signals from the 200 accounts. Given a testing signal $\mathbf{x}$ of an account A, if $\mathbf{x}$ is correctly identified as account A, it is a true positive of account A; if it is wrongly identified as some other account B, it is a false negative of account A and false positive of account B, also counted as a misidentification; if it is not identified as any account, it is counted as a failure of identification. The performance metrics are the average precision of all accounts (Fig. \ref{fig:precision}), the average recall of all accounts (Fig. \ref{fig:recall}), the misidentification rate (total number of misidentification divided by 200$\times$5, Fig. \ref{fig:missrate}), and the failure of identification rate (total number of failure of identification divided by 200$\times$5, Fig. \ref{fig:failrate}). Due to the stochastic nature of neural network, the results are obtained by averaging the performance of five repetitions of the same experiment with the same parameter settings. The ID verification step is not included in this evaluation. These results show that our FMHash framework performs consistently in the user identification task on our dataset with different hash code sizes. In general, longer hash code size provides better security since it is more difficult to guess the hash code without knowing the writing content, but it is also more difficult to train due to the added parameters. Also, a larger fuzziness tolerance $l$ leads to less failure of identification (\textit{i.e.}, improved recall) but more misidentification. In a practical identification system, we recommend to set $l=0$ without ID verification for simplicity or set $l=2$ with ID verification for better security.

\begin{figure}[]
\begin{center}
\includegraphics[width=3.25in]{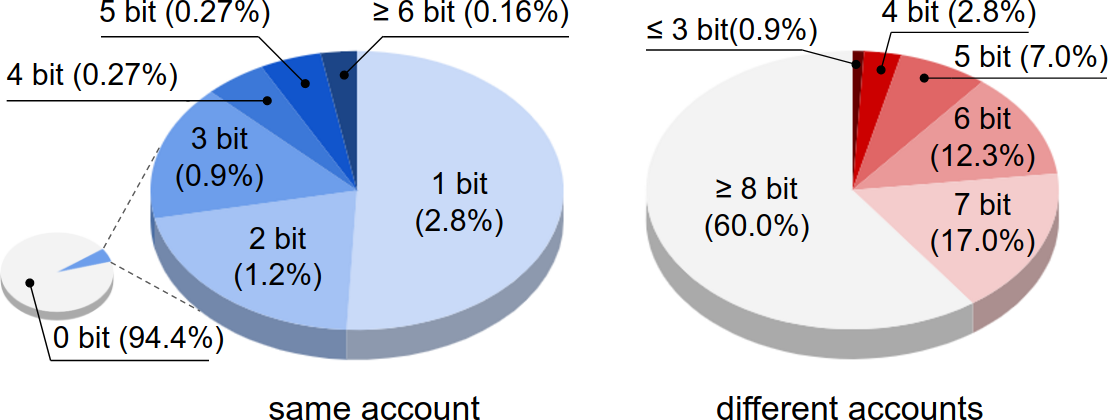}
\end{center}
\vspace{-0.1in}
   \caption{Distribution of the Hamming distance between the account hash code and the hash code of a testing signal for the same account (left) and different accounts (right). The left figure is obtained by counting the Hamming distances of all 200$\times$5 pairs of $\mathbf{b}^{(i)}$ and $\mathbf{b}^{ID}$, where  $\mathbf{b}^{(i)}$ and $\mathbf{b}^{ID}$ are from the same account. The right figure is obtained by counting the Hamming distances of all 200$\times$199$\times$5 pairs of $\mathbf{b}^{(i)}$ and $\mathbf{b}^{ID}$, where $\mathbf{b}^{(i)}$ and $\mathbf{b}^{ID}$ are from different accounts. The hash code size is 16 bits.}
\label{fig:pie}
\vspace{-0.1in}
\end{figure}

Next, we evaluate how much fuzziness is in the hash code caused by the inherent variation in the in-air-handwriting. As shown in Fig. \ref{fig:pie} (left), 1.6\% of the testing signals are hashed more than 2 bits away from the their real accounts. Such fuzziness is mitigated by the separation of the hash codes of different classes, as shown in Fig. \ref{fig:pie} (right), \textit{i.e.}, in 99\% of the case the hash code of a signal of one account is at least three bits far away from the hash code of other accounts.

\begin{figure}[]
\begin{center}
\includegraphics[width=3.25in]{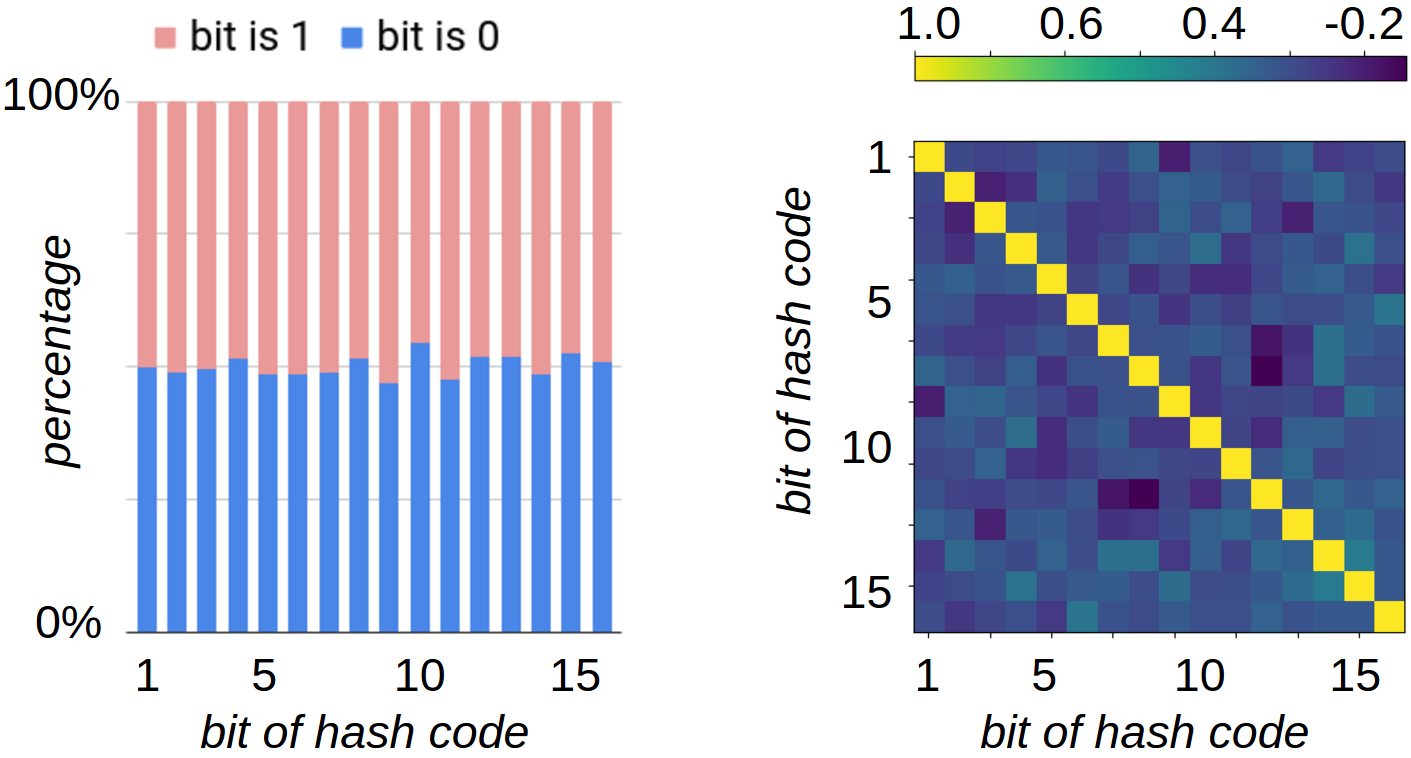}
\end{center}
\vspace{-0.1in}
   \caption{Distribution of zero and one (left) and correlation (right) of each bit in the hash code. The hash code size is 16 bits.}
\label{fig:bit_dist}
\vspace{-0.1in}
\end{figure}

\begin{figure*}
\begin{center}
\includegraphics[width=6in]{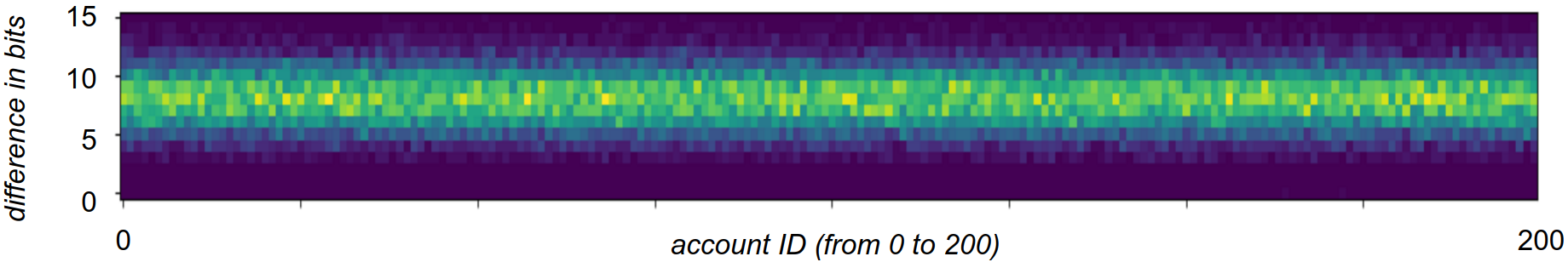}
\vspace{-0.1in}
\end{center}
   \caption{Distribution of the account hash code distance. The column $i$ is the distribution of the Hamming distance $|\mathbf{b}^{(i)} - \mathbf{b}^{(j)}|$ for all accounts where $i \ne j$.}
\label{fig:bit_diff}
\vspace{-0.1in}
\end{figure*}

Then, we study how the hash codes of all accounts are placed in the Hamming space. First, the distribution of zero and one of each bit in the hash code generated in an experiment is shown in Fig. \ref{fig:bit_dist} (left). There are roughly equal amounts of zeros and ones in each bit, indicating that the hash codes are evenly spread. Second, the correlation of every bit pair is shown in Fig. \ref{fig:bit_dist} (right). In this figure, the correlation is close to zero for every pair of bit $i$ and $j$ if $i \ne j$, indicating that each bit carries different information in the hash code. Third, the distribution of the distances of hash codes between any two accounts are shown in Fig. \ref{fig:bit_diff}, where the minimum distance is 3 to 4 bits, the average is 7 to 8 bits, and the maximum is 13 to 14 bits. From this figure, we can see that hash codes of the accounts are sparsely located in the Hamming space and the distance between any two accounts are at least a few bits away. This property of sparsity is the key for an identification system, and it is from our careful design of the regularizer.

At last, we compare our approach with a DSH-like \cite{DSH} regularizer and the commonly used tanh relaxation. The results are shown in Table \ref{tb:compare}. For fair comparison, in the DSH-like approach, the same neural network architecture as FMHashNet is used and only the regularizer is changed to that in DSH. In this method, the regularizer scalar is empirically chosen to be 0.1 (best result achievable), and pair margin $m$ is set to $2B$ as suggested in the original paper. Similarly, in the tanh relaxation approach, the same network architecture is used but only the projection layer is changed to tanh activation and the loss function does not have regularization. In this method, the pair margin $m$ is set to $6$ (separation of at least 3 bits), and the initial learning rate is set to 1e-5 to avoid gradient explosion or vanishing. All compared methods have the same training method, pair selection strategy, and testing protocol. From the results we can see that the compared methods are not optimized to achieve the convergence of hash code with variation in the handwriting and the property of sparsity at the same time. Because the distance between a pair of two accounts are not separated enough, an identification request can get many wrong hash table search results when the bit tolerance increases.

\subsection{Discussions}

We design the FMHashNet as a pure convolutional neural network (CNN) on temporal normalized signals instead of a recurrent neural network (RNN) commonly use in recent signature verification systems \cite{tolosana2018exploring} mainly for speed and simplicity. First, FMHash is essentially a way to input an account ID so it must be able to be retrained or fine-tuned in a few minutes given new accounts are registered (which makes an RNN solution less favorable). Second, we believe that it is difficult to fully learn the long term dependency of handwriting strokes with very limited data for an RNN. Third, to generate a hash code of fixed size representing the whole signal, an RNN needs to keep a large number of hidden states and use them to output the hash code after the last sample of the signal is processed, which further increases the difficulty of training because of the lengthy backpropagation-through-time process.

\section{CONCLUSIONS AND FUTURE WORK}

In this paper, we proposed a user identification framework named FMHash that can generate a compact binary hash code and efficiently locate an account in a database given a piece of in-air-handwriting of an ID string. the empirical results obtained from our prototype evaluation demonstrates the feasibility of the idea of deep hashing of in-air-handwriting for user identification. The ability to convert a finger motion signal to fuzzy hash code gives FMHash great potential for sign-in over gesture input interface. However, it has certain limitations such as the requirement of retraining the neural network on creating new accounts or updating an existing ID. So far, our dataset has limited size and time span. In the future, we will continue improving the proposed framework, investigating the long term performance, and designing possible key generation schemes from the fuzzy hash.

\bibliographystyle{IEEEtran}
\bibliography{reference}

\end{document}